\documentclass[10pt,twocolumn,letterpaper]{article}

\usepackage{iccv}
\usepackage{times}
\usepackage{epsfig}
\usepackage{epstopdf}
\usepackage{graphicx}
\usepackage{amsmath}
\usepackage{amsfonts}
\usepackage{amssymb}
\usepackage{array}
\usepackage{tabularx}
\usepackage{color}
\usepackage{enumitem}
\usepackage{multirow}
\usepackage{multicol}
\usepackage{booktabs}
\usepackage{array,hhline}
\usepackage{algpseudocode}
\usepackage{algorithmicx}
\usepackage{subfigure}
\usepackage{caption}
\usepackage[T1]{fontenc}
\pdfoutput=1
\usepackage{mathrsfs}
\usepackage{bm}
\makeatletter
\newif\if@restonecol
\makeatother

\usepackage[ruled,linesnumbered]{algorithm2e}
\newcommand{\nosemic}{\renewcommand{\@endalgocfline}{\relax}}
\newcommand{\dosemic}{\renewcommand{\@endalgocfline}{\algocf@endline}}
\let\oldnl\nl
\newcommand{\nonl}{\renewcommand{\nl}{\let\nl\oldnl}}

\makeatother

\usepackage[pagebackref=true,breaklinks=true,letterpaper=true,colorlinks,bookmarks=false]{hyperref}

\iccvfinalcopy 

\def\iccvPaperID{1706} 

\ificcvfinal\fi

\begin{document}

\title{Learning from Noisy Labels via Dynamic Loss Thresholding}
\author{
Hao Yang,\textsuperscript{\rm 1,}\textsuperscript{\rm 2,}\footnotemark[1]\ \ 
Youzhi Jin,\textsuperscript{\rm 1,}\textsuperscript{\rm 2,}\footnotemark[1]\ \ 
Ziyin Li,\textsuperscript{\rm 3} 
Deng-Bao Wang,\textsuperscript{\rm 1,}\textsuperscript{\rm 2}
Lei Miao,\textsuperscript{\rm 3} 
Xin Geng,\textsuperscript{\rm 1,}\textsuperscript{\rm 2,}\textsuperscript{\rm 4,}\footnotemark[2]\ \ 
Min-Ling Zhang\textsuperscript{\rm 1,}\textsuperscript{\rm 2,}\textsuperscript{\rm 4,}\footnotemark[2]\\ 
\textsuperscript{\rm 1}\normalsize\textnormal{School of Computer Science and Engineering, Southeast University, Nanjing 210096, China}\\
\textsuperscript{\rm 2}\normalsize\textnormal{Key Laboratory of Computer Network and Information Integration (Southeast University), Ministry of Education, China}\\
\textsuperscript{\rm 3}\normalsize\textnormal{AI Technology Application Department, Huawei Technologies}\\
\textsuperscript{\rm 4}\normalsize\textnormal{Collaborative Innovation Center of Wireless Communications Technology, China}\\
\normalsize\textnormal{\{yang\_h, jinyouzhi, wangdb, xgeng, zhangml\}@seu.edu.cn, \{liziyin, lei.miao\}@huawei.com}
}

\maketitle
\footnotetext[1]{Equal Contribution. This work was done at Huawei Technologies.}
\footnotetext[2]{Correspondence to: Min-Ling Zhang (zhangml@seu.edu.cn) and Xin Geng (xgeng@seu.edu.cn).}
\ificcvfinal\thispagestyle{empty}\fi
\begin{abstract}
    Numerous researches have proved that deep neural networks (DNNs) can fit everything in the end even given data with noisy labels, and result in poor generalization performance. However, recent studies suggest that DNNs tend to gradually memorize the data, moving from correct data to mislabeled data. Inspired by this finding, we propose a novel method named \textit{Dynamic Loss Thresholding} (DLT). During the training process, DLT records the loss value of each sample and calculates dynamic loss thresholds. Specifically, DLT compares the loss value of each sample with the current loss threshold. Samples with smaller losses can be considered as clean samples with higher probability and vice versa. Then, DLT discards the potentially corrupted labels and further leverages supervised learning techniques. Experiments on CIFAR-10/100 and Clothing1M demonstrate substantial improvements over recent state-of-the-art methods. 
    
    In addition, we investigate two real-world problems for the first time. Firstly, we propose a novel approach to estimate the noise rates of datasets based on the loss difference between the early and late training stages of DNNs. Secondly, we explore the effect of hard samples (which are difficult to be distinguished) on the process of learning from noisy labels.
\end{abstract}

\begin{figure}[htbp]
\centering
\subfigure[\!]{ 
\begin{minipage}{0.48\columnwidth}
\centering 
\includegraphics[scale=0.30,trim=40 -10 -30 0]{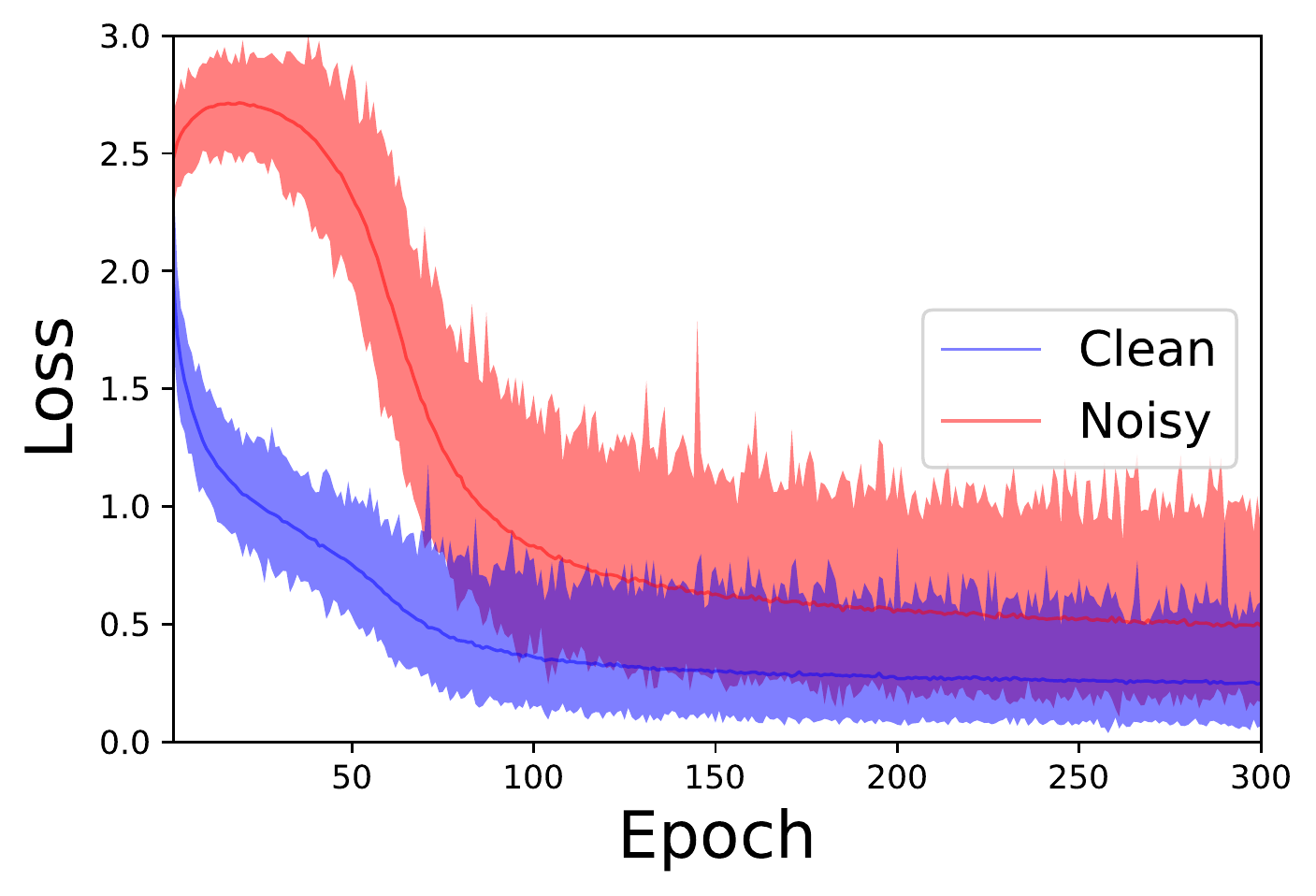}
\end{minipage}}
\subfigure[\!\!\!\!\!\!\!\!\!\!\!]{
\begin{minipage}{0.48\columnwidth}
\centering 
\includegraphics[scale=0.30, trim=20 -10 0 0]{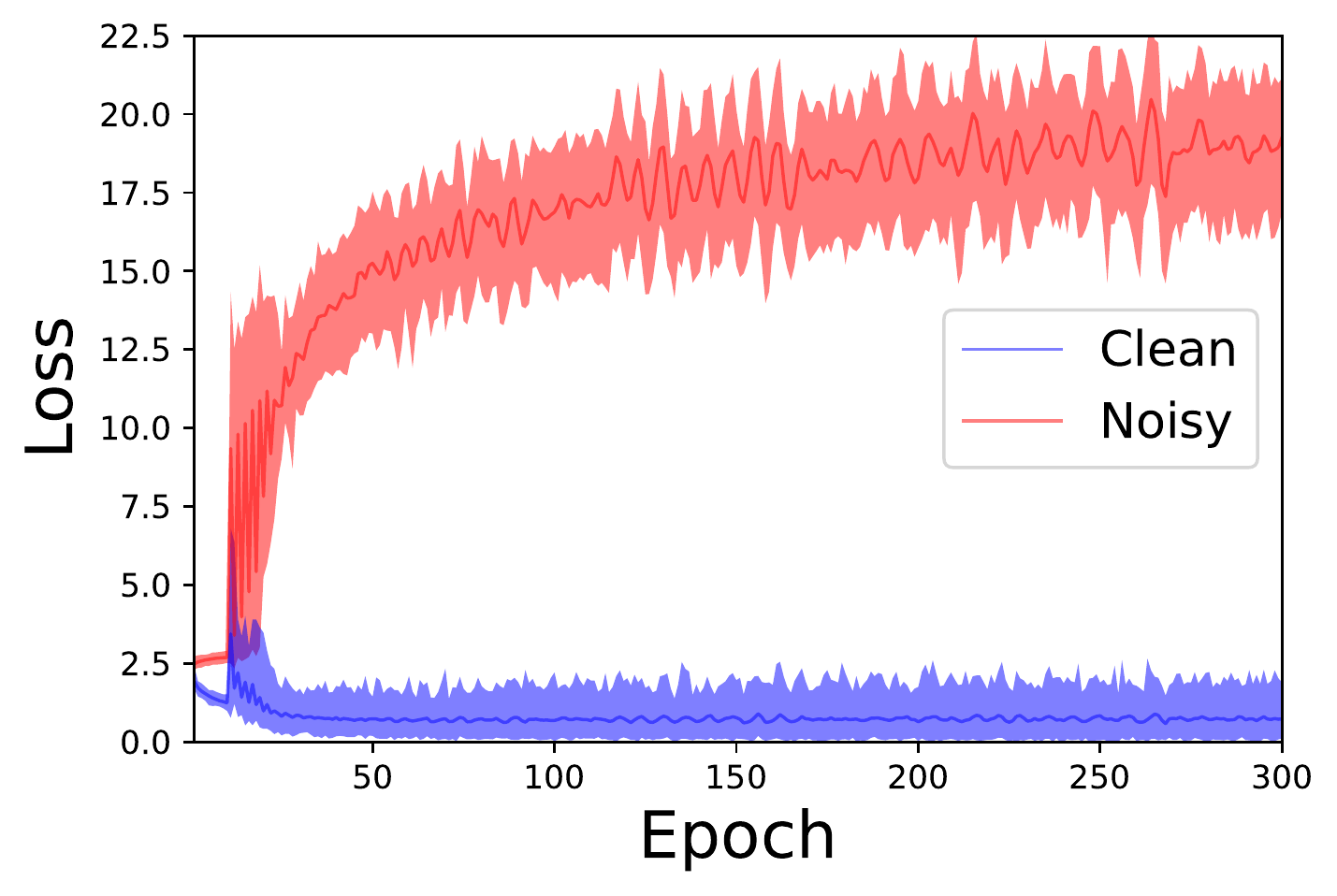}
\end{minipage}}
\vspace{-5pt}
\caption{Cross-entropy loss on CIFAR-10 under 50\% symmetric label noise. (a) Training with cross-entropy loss results in fitting the noisy labels. (b) Using DLT avoids fitting label noise. The bold lines represent the means of losses and the shaded areas are the ranges of all samples.} 
\vspace{-10pt}
\end{figure}

\section{Introduction}

Although deep neural networks (DNNs) have achieved great success for image classification tasks~\cite{he2016deep,krizhevsky2012imagenet}, their excellent performance mainly relies on large-scale datasets with clean label annotations. However, it is extremely expensive and time-consuming to label high-quality datasets, thus deep models are usually trained on data with lots of corrupted labels. As a result, dealing with label noise is a common adverse scenario which requires attention and has been extensively studied these years~\cite{arazo2019unsupervised, han2018co, jiang2018mentornet, li2020dividemix}.

A recent study on the generalization capabilities of deep networks~\cite{zhang2016understanding} demonstrates that DNNs can easily overfit to noisy labels and result in poor generalization performance. However, even though deep networks can fit everything in the end, they \textit{learn patterns first}~\cite{arpit2017closer}, and this suggests that DNNs gradually memorize the data, moving from correct data to mislabeled data. As shown in Figure 1(a), DNNs fit the correctly labeled samples (clean samples) before fitting noisy samples, resulting in notably larger loss values for noisy samples in the early training stage.

Existing methods for learning from noisy labels can be grouped into three main categories. The first one is based on label correction which aims to correct noisy labels to the ground-truth ones~\cite{li2017learning, tanaka2018joint, vahdat2017toward}. For example, the recent proposed \textit{PENCIL}~\cite{yi2019probabilistic} utilizes back-propagation to correct image labels and update the network parameters simultaneously in an end-to-end manner. The second one is based on the robust loss function~\cite{ma2020normalized, wang2019symmetric, zhang2018generalized}. Ghoshet al.~\cite{ghosh2017robust} proves that the loss functions which satisfy the symmetric condition, such as Mean Absolute Error (MAE), would be inherently tolerant to both uniform and class conditional label noise. The third one is based on sample selection which involves selecting correctly labeled samples from a noisy training dataset
~\cite{arazo2019unsupervised, huang2019o2u, li2020dividemix}. \textit{Co-teaching}~\cite{han2018co, yu2019does} is a representative framework on this line which trains two networks where each network selects small-loss samples to teach another one.

In light of these recent advances, we propose \textit{dynamic loss thresholding} (DLT) to avoid fitting noisy labels when training deep models, as shown in Figure 1(b). Specifically, DLT records the loss value of each sample during training and calculates loss thresholds dynamically. By comparing the loss value of each training sample with the current loss threshold, we expect to distinguish the potentially clean and noisy samples during training. Samples with smaller losses can be considered as correctly labeled samples and vice versa. Then, DLT discards these potentially corrupted labels and treats the corresponding samples as unlabeled data, thus we can leverage semi-supervised learning techniques like \textit{Mixup}~\cite{zhang2017mixup} to improve the performance. Experiments on various benchmarks demonstrate substantial improvements over recent state-of-the-art methods. Furthermore, DLT is also of excellent generalization and flexibility and we empirically demonstrate that our noisy label detection method is still effective when combining with other methods.

In addition, the noise rates of datasets are always unknown but crucial to many real-world situations due to the fact that numbers of existing methods need the noise rate as their prior knowledge. 
In this paper, we further propose a novel method to estimate the noise rate based on the patterns of losses. In particular, our method calculates the \textit{loss difference} between the early and late deep network training stages and dynamically fits a Gaussian Mixture Model (GMM) on per-sample loss difference to divide the training samples into a clean set and a noisy set. Accordingly, noise rate can be obtained by calculating the proportion of clean samples to total samples.

In real-world situations, there exists a common class of samples which are always quite close to the decision boundary and hence difficult to be distinguished by DNNs. Intuitively, we call these samples \textit{hard samples}. Hard samples can be defined as a subset of clean samples but easy to be mistaken for other classes. However, nearly no research discussed hard samples in the field of learning from noisy labels to our knowledge. By means of our proposed method, we investigate the effect of hard samples during deep network training for the first time.
In summary, our main contributions are as follows:
\begin{itemize}
\item We propose a novel noisy label detection method DLT, which is based on \textit{dynamic loss thresholds}. Combined with semi-supervised learning techniques, our whole framework achieves state-of-the-art performance. We experimentally show that DLT is of excellent generalization and flexibility.
\item We provide a method to estimate the noise rates of datasets based on \textit{loss difference}.
This is a key complement to eliminate the dependence on using noise rate as common prior knowledge.
\item We provide insights into the effect of hard samples on the training of DNNs. We do this for not only clarifying the effectiveness of our method, but also finding out the patterns of hard samples when learning from noisy labels.
\end{itemize}

\section{Related Work}
\subsection{Learning from Noisy Labels}
Existing approaches for training DNNs from noisy labels can be roughly divided into three categories: 1) label correction methods, 2) loss correction methods, and 3) sample selection methods.

\vspace{5pt}
\textbf{Label correction} This family of methods aims to relabel the corrupted labels. One research line tries to formulate explicit or implicit noise models to characterize the distribution of noisy and true labels using directed graphical models~\cite{xiao2015learning} , Conditional Random Fields~\cite{vahdat2017toward} , knowledge graph~\cite{li2017learning}, or neural networks~\cite{lee2018cleannet, veit2017learning}. However, to recover the ground-truth labels, these approaches usually require the support from a small set of clean samples. Recently, Tanaka et al.~\cite{tanaka2018joint} propose a method which relabels samples using network predictions by alternately updating network parameters and labels. \textit{PENCIL}~\cite{yi2019probabilistic} updates both network parameters and label
estimations in an end-to-end manner and does not need an auxiliary clean dataset.

\vspace{5pt}
\textbf{Loss correction} Another line of learning from noisy labels seeks to modify the loss function to achieve robustness. Patrini et al.~\cite{patrini2017making} propose a loss correction method based on pre-calculated \textit{Backward} or \textit{Forward} noise transition matrix. Goldberger and Ben-Reuven~\cite{goldberger2016training} propose to augment the correction architecture by adding an additional linear layer on top of the neural network. Besides, Ghosh et al.~\cite{ghosh2017robust} prove that for multi-class classification, the loss functions which satisfy the symmetric condition, such as Mean Absolute Error (MAE), would be inherently tolerant to both uniform and class conditional label noise. However, Zhang and Sabuncu~\cite{zhang2018generalized} show that it is not able to achieve good performance by learning DNNs with MAE due to slow convergence caused by gradient saturation. Based on these findings, Zhang and Sabuncu~\cite{zhang2018generalized} propose \textit{Generalized Cross Entropy} (GCE) which applies a negative Box-Cox transformation. \textit{Symmetric Cross Entropy} (SCE)~\cite{wang2019symmetric} combines Reverse Cross Entropy (RCE) (which satisfies the symmetric condition) together with the Cross Entropy loss. \textit{Active Passive Loss} (APL)~\cite{ma2020normalized} combines two robust loss functions namely active loss and passive loss which mutually boost each other.

\vspace{5pt}
\textbf{Sample selection} Besides the above two kinds of methods, numerous studies involve selecting potentially clean samples from a noisy training dataset. \textit{Decouple}~\cite{malach2017decoupling} trains the model using selected samples based on the discrepancy between the two classifiers. \textit{MentorNet} introduces a data-driven curriculum learning paradigm in which a pre-trained mentor network guides the training of a student network. \textit{Co-teaching}~\cite{han2018co} trains two DNNs simultaneously, and let them teach each other with some selected samples during every mini-batch. \textit{O2U-Net}~\cite{huang2019o2u} selects samples with small losses and keeps the status of the learned network in transferring from overfitting to underfitting cyclically. A recent study~\cite{li2020dividemix} proposed a method named \textit{DivideMix} which trains two networks simultaneously and fits a Gaussian Mixture Model (GMM) on its per-sample loss distribution to divide the training samples into a labeled set and an unlabeled set. By leveraging the semi-supervised learning technique \textit{MixMatch}~\cite{berthelot2019mixmatch}, they achieved state-of-the-art performance on several benchmarks.

\subsection{Semi-Supervised Learning}
Semi-Supervised learning (SSL) methods are used to learn in the presence of both labeled and unlabeled data, which is naturally applicable to the problems of learning from noisy labels after discarding the labels of potentially mislabeled samples. Current SSL methods can be broadly divided into the following categories: \textbf{consistency regularization}~\cite{laine2016temporal, park2018adversarial, rasmus2015semi} which is based on the assumption that if a realistic perturbation was applied to the unlabeled data, the prediction should not be changed significantly, \textbf{proxy-label methods}~\cite{iscen2019label, shi2018transductive} which leverage a pre-trained model on the labeled samples to produce additional training samples by labeling unlabeled samples, \textbf{generative models methods}~\cite{kingma2014semi, springenberg2015unsupervised} which model the real data distribution from the training dataset and then generate synthetic samples as augmentations. An emerging line of work is a set of holistic approaches that try to combine different dominant methods in SSL~\cite{berthelot2019remixmatch, berthelot2019mixmatch, sohn2020fixmatch}. Moreover, a learning principle, namely \textit{Mixup}~\cite{zhang2017mixup}, is usually employed in those hybrid methods.

\section{Methodology}
We begin by introducing our dynamic loss threshold based method DLT. Besides, a novel approach is proposed to estimate the noise rates of datasets through loss difference. Lastly, we introduce two strategies to generate hard samples.

\begin{figure}[htbp]
\centering
\subfigure[\!\!\!\!\!\!\!\!\!\!]{ 
\begin{minipage}{0.8\columnwidth}
\centering 
\includegraphics[trim=150 250 180 180,clip,width=\textwidth]{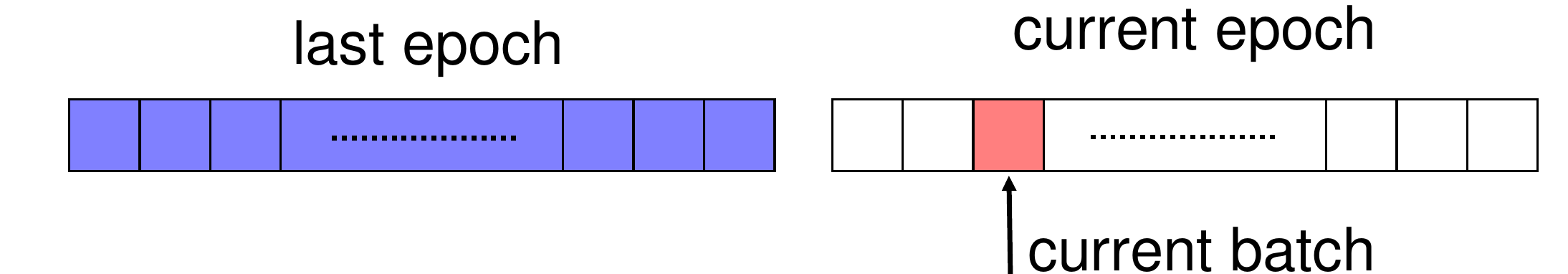}
\end{minipage}}
\quad
\subfigure[\!\!\!\!\!\!\!\!\!\!\!\!]{
\begin{minipage}{0.8\columnwidth}
\centering 
\includegraphics[trim=150 80 180 350,clip,width=\textwidth]{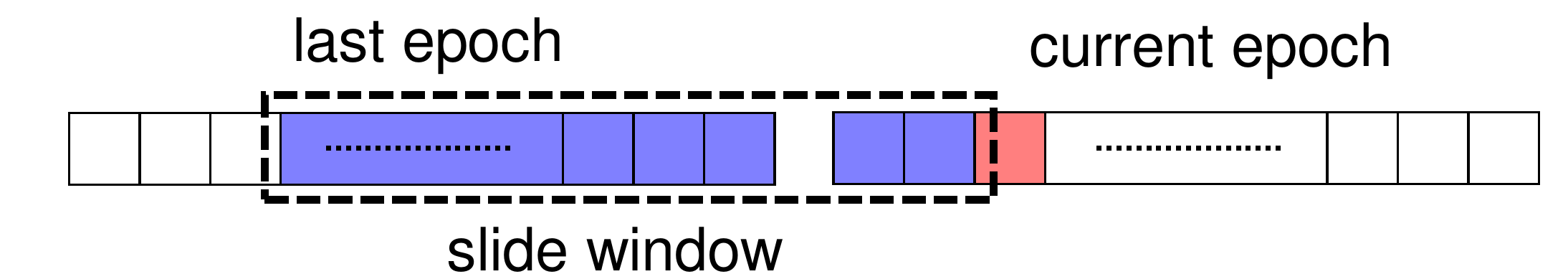}
\end{minipage}}
\vspace{-5pt}
\caption{(a) Last epoch strategy. (b) Slide window strategy. The purple windows represent the batches which are utilized to calculate dynamic loss thresholds.} 
\vspace{-10pt}
\end{figure}

\subsection{The framework of DLT}
Formally, let $\mathcal{D} =(\mathcal{X,Y} )=\left \{ \left ( x_{i}, y_{i}  \right )  \right \} _{i=1}^{N} $ denote the training data, where $x_i$ is an image and $y_i\in \left \{ 0,1 \right \} ^{C} $ is the one-hot label over $C$ classes. The cross-entropy loss for sample $x_i$ is:
\begin{equation}
\begin{aligned}
l_i=-\sum_{c=1}^{C}y_{i}^{c}\log_{}(P_{model}^{ c}(x_i))
\end{aligned}
\end{equation}
where $P_{model}^{ c}$ is the deep model's softmax output for class $c$. Let $b$ denote the size of each mini-batch and $m = \lceil N / b\rceil $ denote the number of batches in each epoch. Let $\mathcal{B}_p^t$ denote the $p_{th}$ batch at the $t_{th}$ epoch, then the loss values of the samples in $\mathcal{B}_p^t$ can be recorded as $ \mathcal{L}_p^t=\left \{l_{p,1}^{t},l_{p,2}^{t},...,l_{p,b}^{t}\right\} $. Let $w$ be the noise rate and $q$ be the selection proportion for training, which is set as $q=1-w$. Besides, $Quantile(\mathcal{L}_p^{t}, q)$
represents the function to calculate the quantile that separates $\mathcal{L}_p^{t}$.

In general, DLT can be divided into two steps: noisy label detection based on dynamic loss thresholds and then exploit both potentially clean and noisy samples along with SSL techniques.
\subsubsection{Noise detection by dynamic loss thresholds}

\quad As we described above, DNNs usually fit the clean samples before fitting noisy samples, resulting in notably larger loss values for noisy samples in the early training stage. Therefore, a straight strategy is to identify noisy labels based on a specific loss threshold. Here we introduce two strategies to obtain dynamic loss thresholds.

\vspace{5pt}
\textbf{Last epoch strategy} When training on current batch $\mathcal{B}_p^t$, the loss thresholds can be calculated from the recorded loss values of the last epoch, i.e. $\mathcal{L}^{t-1} = \mathcal{L}_1^{t-1}\cup \mathcal{L}_2^{t-1},..., \mathcal{L}_m^{t-1}$, as shown in Figure 2(a). Notably, all samples in current epoch are compared with the same loss threshold when using this strategy. Based on selection proportion $q$, the loss threshold $\tau_p^{t}$ can be calculated as:
\begin{equation}
\begin{aligned}
\tau_p^{t} =Quantile(\mathcal{L}^{t-1}, q)
\end{aligned}
\end{equation}

\vspace{5pt}
\textbf{Slide window strategy} Let $s$ denote the length of slide window. As shown in Figure 2(b), when training on current batch $\mathcal{B}_p^t$, loss thresholds can be calculated from the last $s$ batches. We denote $\mathcal{L}_{[-s:]}$ as the set that consists of the loss values of samples in the last $s$ batches. Notably, this slide window strategy reflects the learned model more timely. Based on selection proportion $q$, $\tau_p^{t}$ can be calculated as:
\begin{equation}
\begin{aligned}
\tau_p^{t} =Quantile( \mathcal{L}_{[-s:]}, q)
\end{aligned}
\end{equation}
Then, for each sample $x_i$, DLT compares its loss value $l_{i}$ with the loss threshold $\tau_p^{t}$: $x_i$ is considered as true-labeled if $l_{i} \le \tau _p^{t}$, otherwise it might be mislabeled. Accordingly, training data $\mathcal X$ can be divided into two parts: $\mathcal X_{clean}$ and $\mathcal X_{noisy}$, representing the potentially clean and noisy set respectively. Besides, we use an illustrative experiment to verify the effectiveness of DLT. As shown in Figure 3, most of the clean samples are retained during training, while few noisy samples are infiltrated in the training process.

\begin{figure}[t]
\begin{center}
\includegraphics[trim=0 0 0 0,clip, width=0.9\linewidth]{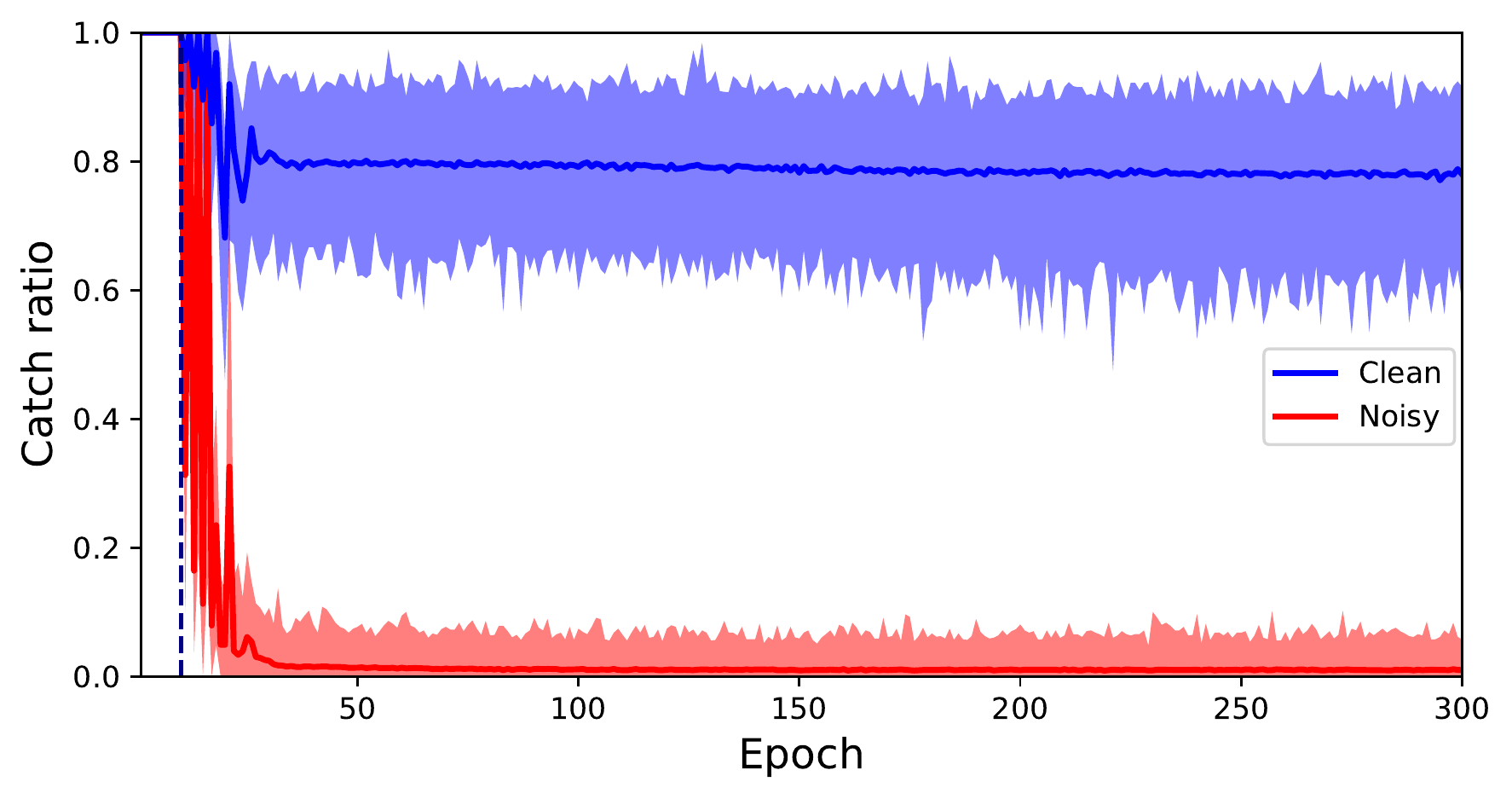}
\end{center}
\vspace{-10pt}
\caption{The ratios of clean and noisy samples which are sent to deep model for training under 0.5 symmetric noise.}
\label{fig:long}
\label{fig:onecol}
\end{figure}
\begin{figure}[t]
\begin{center}
\includegraphics[trim=0 0 15 10,clip, width=0.98\linewidth]{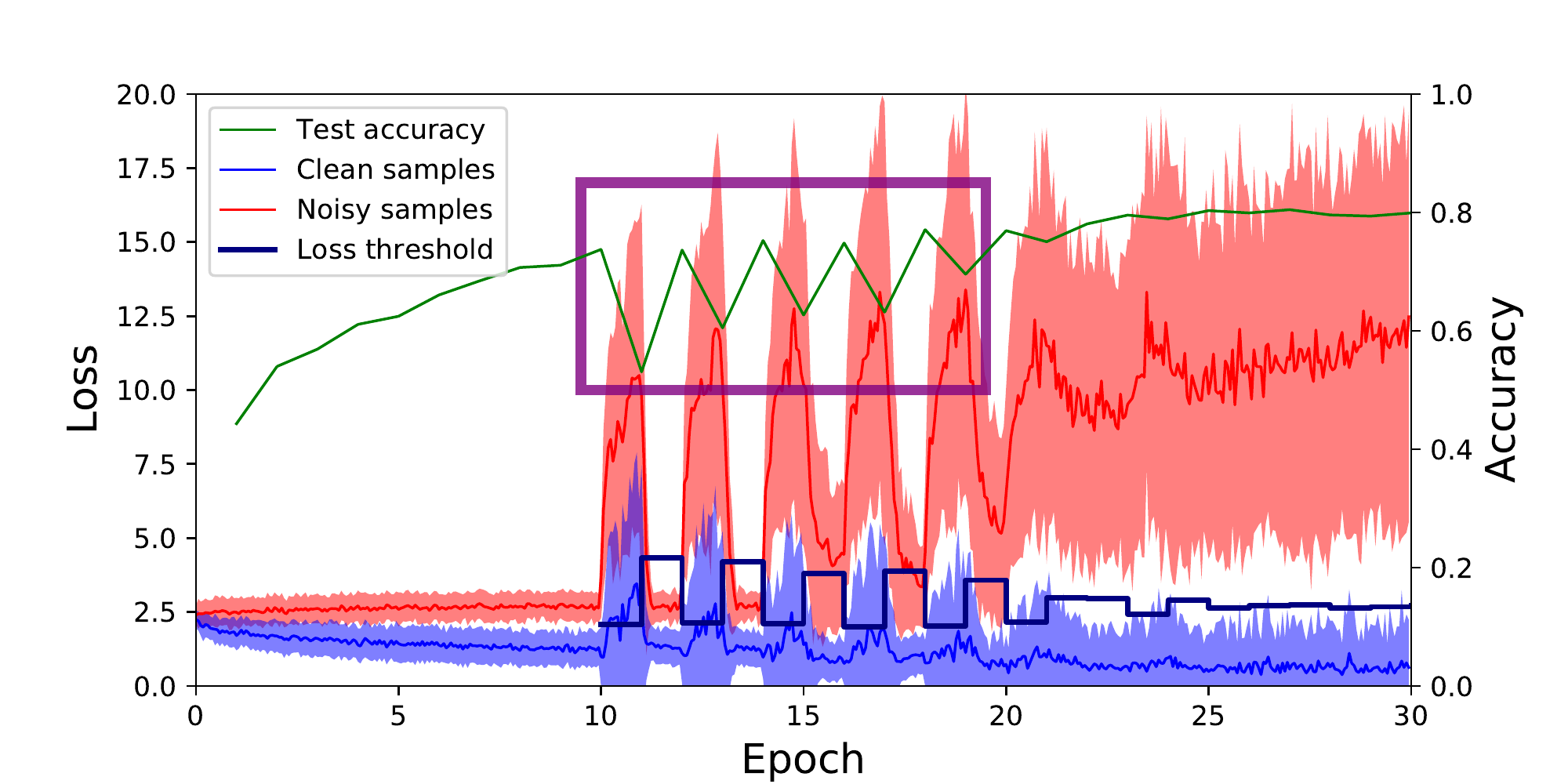}
\end{center}
\vspace{-20pt}
\caption{Test accuracy fluctuates in the early stage of training (warm up for 10 epochs).}
\vspace{-10pt}
\end{figure}
\subsubsection{Utilizing semi-supervised learning techniques}

\quad After obtaining $\mathcal{X}_{clean}$ and $\mathcal{X}_{noisy}$, SSL techniques are leveraged in the following steps. Firstly, as is typical in many SSL methods, we use data augmentation on both clean and noisy data for $K$ times. Secondly, clean and noisy samples are utilized respectively: 1) for $\mathcal{X}_{clean}$, cross-entropy loss and \textit{mixup} are leveraged; 2) for $\mathcal{X}_{noisy}$, its original corrupted labels will be discarded and our model will generate pseudo-labels $\mathcal{Y}_{pseudo}$ for them. Then a sharpening function is applied on $\mathcal{Y}_{pseudo}$:
\vspace{-5pt}
\begin{equation}
\begin{aligned}
\hat{\mathcal{Y}}_{pseudo}^c = {\mathcal{Y}_{pseudo}^{c}}^{\frac{1}{T}} \bigg/\sum_{k=1}^{C} {\mathcal{Y}_{pseudo}^{k}}^{\frac{1}{T}}, \ \ c=1,2,...,C
\end{aligned}
\end{equation}
where $T$ denotes the sharpening temperature. According to the work in ~\cite{arazo2020pseudo}, in order to utilize the potential information from noisy samples effectively, we randomly choose a part of samples in $\mathcal X_{clean}$ to \textit{Mixup} with $\mathcal X_{noisy}$. Moreover, mean squared error is employed for $\mathcal X_{noisy}$. Lastly, to avoid assigning all samples to a single class, we apply the regularization used by ~\cite{arazo2019unsupervised, tanaka2018joint}, which uses a uniform prior distribution $\pi$ to regularize the model output in the mini-batch:
\begin{equation}
\begin{aligned}
\mathcal{L}_{reg} = \sum_{c=1}^{C} \pi_c log\left(\pi_c\bigg/\frac{1}{|\mathcal{X}|}\sum_{i=1}^{N}P_{model}^c(x_i)\right) 
\end{aligned}
\end{equation}
where $\pi_c=1/C$. In summary, the loss $\mathcal L_{clean}$ on $\mathcal X_{clean}$ is the cross-entropy loss and the loss $\mathcal L_{noisy}$ on $\mathcal X_{noisy}$ is the mean squared error. Along with the regularization term $\mathcal{L}_{reg}$, the total loss of DLT is: 
\begin{equation}
\begin{aligned}
\mathcal L = \mathcal L_{clean} + \lambda _{n} \cdot \mathcal L_{noisy} + \lambda _{r} \cdot \mathcal{L}_{reg}
\end{aligned}
\end{equation}
In our experiments, we set $\lambda _{r}$ as $1$ and use $\lambda _{n}$ to control the contribution of noisy samples.

\subsubsection{Leveraging decreasing loss thresholds}
\quad For initial convergence of the algorithm, DLT needs to \textit{warm up} the model in the beginning for a few epochs by training on all data using standard cross-entropy loss. As Figure 4 illustrates, the test accuracy of DLT always fluctuates in the early stage of training. The observed fluctuation is caused by that DLT will receive data with different distribution after the warm-up stage, which results in poor generalization performance. DLT leverages a decreasing loss threshold strategy to solve this dilemma. Specifically, the loss threshold is set to decrease gradually as training goes on. Formally, let $T_{warm}$ denote the number of warm-up epochs and $T_{grad}$ denote the number of loss threshold dynamically decreasing epochs. At the $t_{th}$ epoch, we can get the selection proportion $q$ as: 

\begin{equation}
\begin{aligned}
q= \left\{\begin{matrix}
  100\% ,& t \in [1, T_{warm}] , \\
1-{w}\cdot\frac{t-T_{warm}}{T_{grad}} ,& t\in (T_{warm} ,T_{warm}+T_{grad}] ,\\
  1-w ,& otherwise.\\
\end{matrix}
\right.
\end{aligned}
\vspace{5pt}
\end{equation}



\setlength{\textfloatsep}{12pt}
\begin{algorithm}[t]  
    \setlength{\belowcaptionskip}{-10pt}
    \caption{Learning via DLT.}
    \KwIn{training set $\mathcal{D}$, mode $M$, total epochs $T_{total}$, $T$, $w$, $s$, $m$, $T_{warm}$, $T_{grad}$, $\lambda_{u}$, $K$.}
    \KwOut{Model parameters $\bm\theta$.}
    \For{$t=1 \ldots  T_{total}$}
    {
      \For{$p=1 \ldots m$}
      {
        Fetch mini-batch ${\mathcal{B}}$ from ${\mathcal{D}}$;\\
        Calculate loss by $\bm l=\mathcal{L}_{ce}(\mathcal{B},\bm\theta))$;\\
        \If{$t \le T_{warm}$}
        {   \vspace{-10pt}
            \nonl\qquad\qquad\qquad\qquad\qquad\ \ \ \ \  \,\textcolor[rgb]{0.4,0.25,0.15}{$\triangleright$ Warm up}\\
            Update $\bm\theta=SGD(\bm l,\bm\theta)$;
        }
        \Else
        {
            Obtain $q$ using Eq. (7); \\
            \If {$M=LAST \ \ EPOCH$} 
            {
            \vspace{-5pt}
            \nonl\qquad\qquad\qquad\qquad\qquad\qquad\qquad\ \ \ \ \, \textcolor[rgb]{0.4,0.25,0.15}{\qquad\qquad\qquad\ \  $\triangleright$ Last Epoch mode}\\
            Obtain $\tau_p^t$ using Eq. (2); \\
            }
            \Else 
            {\vspace{-10pt}
            \nonl\nonl\qquad\qquad\ \ \ \ \ \ \ \  \,\textcolor[rgb]{0.4,0.25,0.15}{$\triangleright$ Slide Window mode}\\
            Obtain $\tau_p^t$ using Eq. (3); \\
            }
            $\hat{\mathcal{B}}=$Augment($\mathcal{B}$, $K$);\\
            Obtain $\mathcal{B}_{clean}=\{\hat{\mathcal{B}}_i|l_i \le \tau_b^t, l_i \in \bm l \}$;\\
            Obtain $\mathcal{B}_{noisy}=\{\hat{\mathcal{B}}_i|l_i > \tau_b^t, l_i \in \bm l \}$;\\
            Calculate $\mathcal{Y}_{pseudo} = P_{model}(\mathcal{B}_{noisy};\bm\theta)$;\\
            \vspace{-2pt}
            Obtain $\hat{\mathcal{Y}}_{pseudo}$ using Eq. (4);\\
            Calculate $\mathcal{L}_{clean}$, $\mathcal{L}_{noisy}$ and $\mathcal{L}_{reg}$;\\
            $\mathcal{L}=\mathcal{L}_{clean} + \lambda_{n} \cdot \mathcal{L}_{noisy} + \lambda_{r} \cdot \mathcal{L}_{reg}$;\\
            Update $\bm \theta = SGD(\mathcal{L},\bm \theta)$;\\
        }
      }
      Preserve the model outputs of the current epoch;
    }
    \textbf{return} $\bm \theta$.
\end{algorithm}

DLT is summarized in Algorithm 1. The warm-up stage is essential in the beginning phase, since the divergence of clean and noisy samples will highlight in this stage. Then, DLT leverages dynamically decreasing loss thresholds to divide training samples. Lastly, our model generates pseudo-labels for noisy samples and leverages SSL techniques.

\subsection{Estimating noise rate by loss difference }
Based on the discrepancy between loss values of clean and noisy samples, we propose a novel method to estimate the noise rates of datasets. As observed in Figure 5, loss values of clean and noisy samples are significantly discrepant in the early stage. In contrast, loss values of them are similar in the late stage of training. As a result, noisy samples can be detected by calculating the loss difference between the early and late stage. Intuitively, samples with larger loss differences can be considered as noisy samples. Specifically, we conduct a GMM with two components using Expectation Maximization algorithm. For each sample $x_i$, our model calculates the clean probability $\phi _i$, which is the posterior probability $p(g|x_i)$ and $g$ is the Gaussian component with smaller mean (smaller loss). Next, we set a threshold $\theta$ on $\phi _i$ and $\mathcal X_{clean} = \left \{ x_i | \phi_i <\theta \right \} $. Therefore, the number of clean samples $n_{clean} = | \mathcal X_{clean}|$ is obtained. The noise rate $r$ can be estimated as: 
\begin{equation}
\begin{aligned}
r = \frac{n_{clean}}{n} \times 100\%
\end{aligned}
\end{equation}

\begin{figure}[t]
\begin{center}
\includegraphics[trim=0 0 0 0,clip, width=0.8\linewidth]{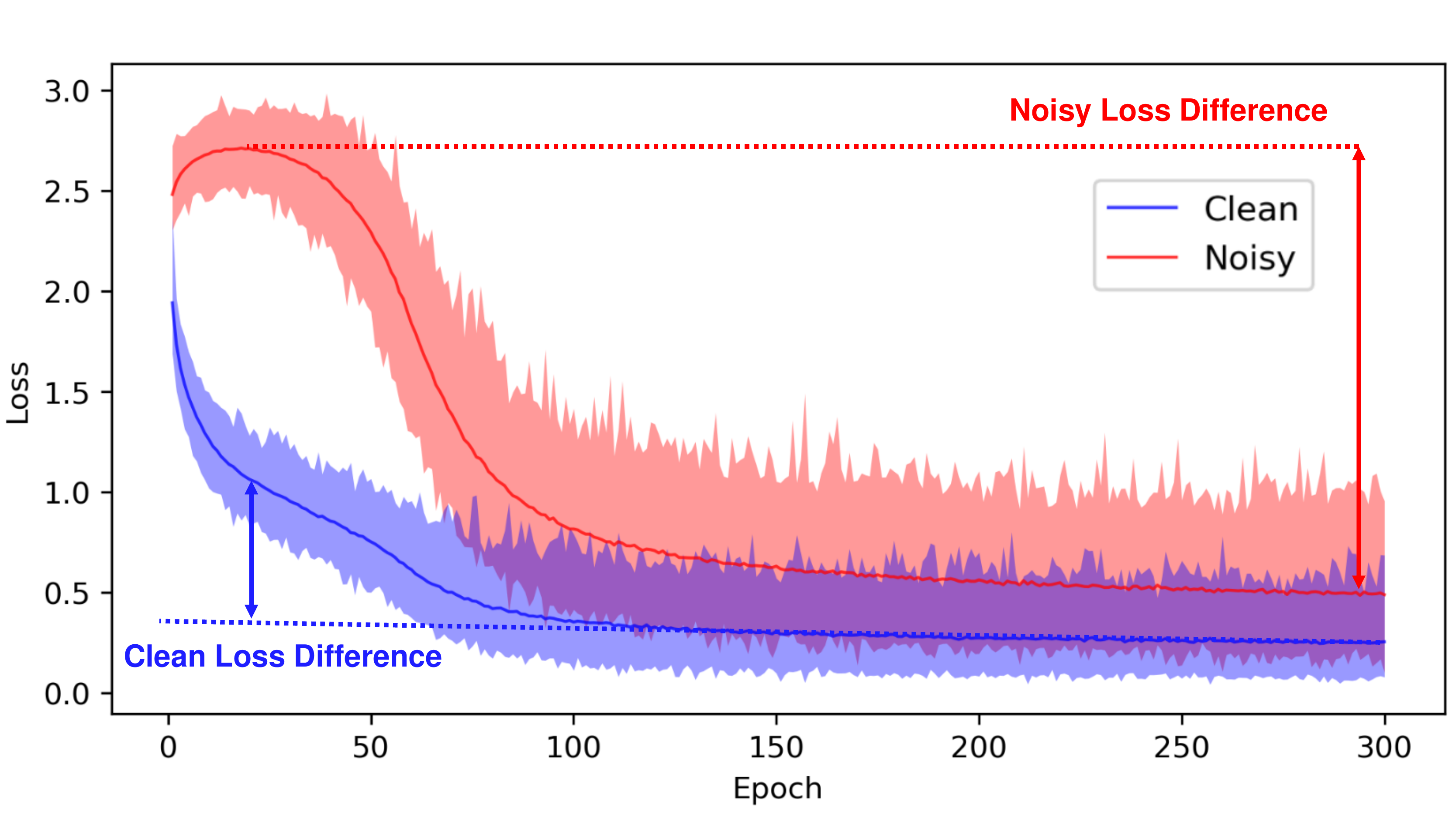}
\end{center}
\vspace{-8pt}
\caption{Loss difference between early and late deep network training stages.}
\label{fig:long}
\label{fig:onecol}
\end{figure}

\begin{figure}[t]
\begin{center}
\hspace{12pt}
\includegraphics[trim=0 100 0 60,clip, width=0.8\linewidth]{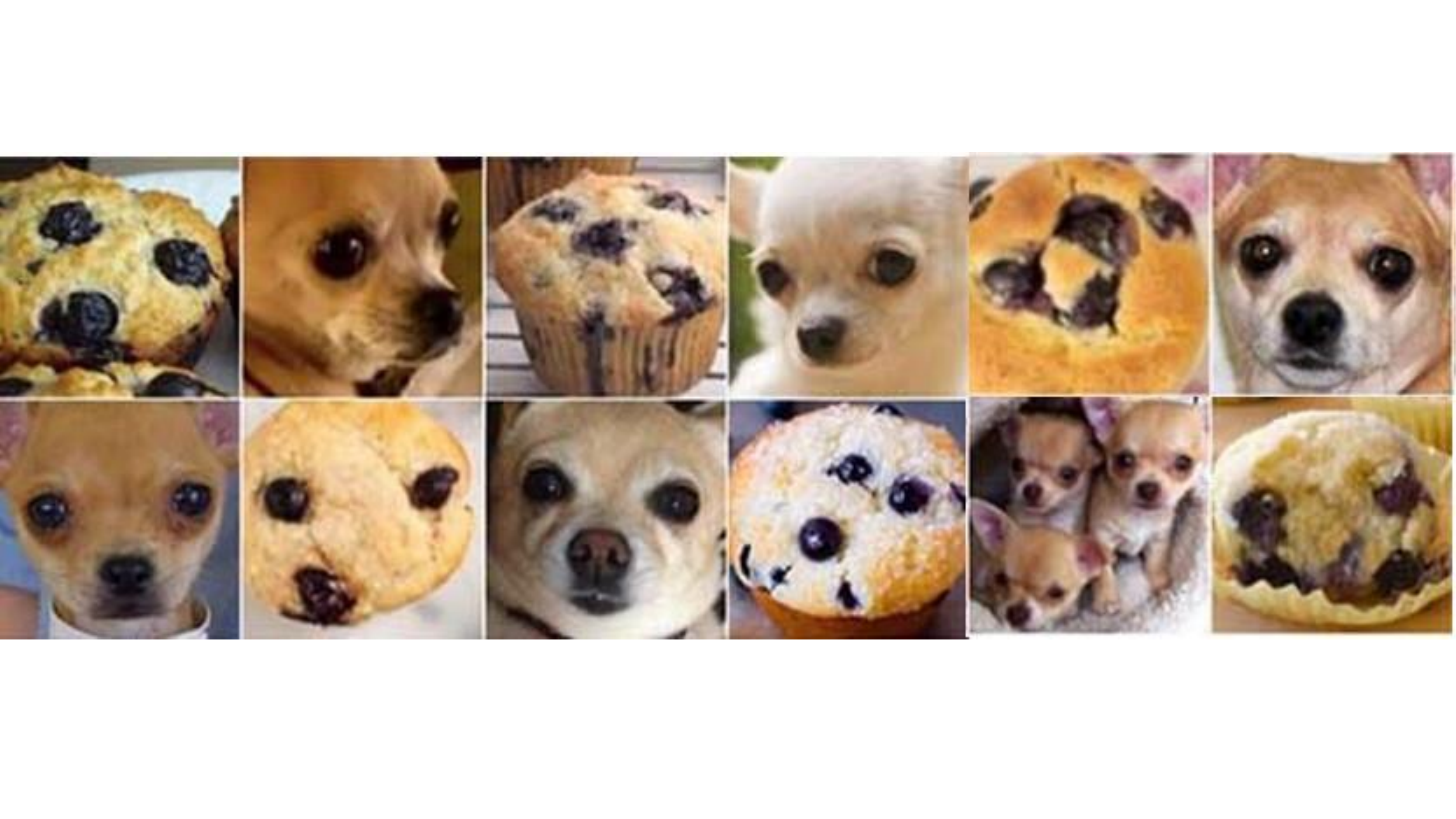}
\end{center}
\vspace{-5pt}
\caption{Examples of hard samples (muffin or dog?).}
\label{fig:long}
\label{fig:onecol}
\vspace{10pt}
\end{figure}

\begin{table*}[t]
\begin{center}
\renewcommand\arraystretch{1.1}
\begin{tabular}{p{3.5cm}<{\centering}p{1cm}<{\centering}|p{1cm}<{\centering}|p{1cm}<{\centering}|p{1cm}<{\centering}||p{1.0cm}<{\centering}|p{1cm}<{\centering}|p{1cm}<{\centering}}
\toprule[1.5pt]\hline
\multicolumn{2}{c|}{Noise Type}&\multicolumn{3}{c||}{Symmetric Noise} & \multicolumn{3}{c}{Asymmetric Noise}\\
\hline
\multicolumn{2}{c|}{Methods / Noise Rate}&0.2& 0.4& 0.6& 0.2& 0.3& 0.4\\
\hline\hline
\multirow{2}{*}{Standard Cross-Entropy}
                &Best& 88.1 & 84.3 & 77.7 & 90.6 & 89.4 & 85.0 \\ 
                &Last& 81.7 & 65.0 & 44.9 & 87.8 & 83.4 & 72.3 \\ \hline
\multirow{2}{*}{APL (2020)}
                &Best& 87.8 & 84.4 & 76.8 & 87.2 & 85.7 & 83.6   \\ 
                &Last& 87.5 & 84.0 & 76.4 & 87.0 & 84.6 & 82.2   \\ \hline
\multirow{2}{*}{SCE (2019)}
                &Best& 92.9 & 90.9 & 87.0 & 92.3 & 91.2 & 87.9 \\ 
                &Last& 90.0 & 81.7 & 62.7 & 90.1 & 85.5 & 79.9 \\ \hline
\multirow{2}{*}{M-correction (2019)}
                &Best& 94.0 & 93.3 & 90.1 & --- & --- & --- \\ 
                &Last& 93.6 & 93.1 & 89.7 & --- & --- & --- \\ \hline
\multirow{2}{*}{DivideMix (2020)}
                &Best& 95.6 & 93.9 & 93.7 & 94.2 & 93.3 & 92.7 \\ 
                &Last& 95.2 & 93.6 & 93.0 & 93.4 & 92.5 & 91.5 \\ \hline
\multirow{2}{*}{DLT (Last Epoch)}
                &Best& 95.7 & \textbf{95.6} & \textbf{94.8} & \textbf{94.4} & \textbf{93.5} & 92.5 \\ 
                &Last& 95.4 & 94.8 & \textbf{94.5} & \textbf{93.9} & 93.0 & 91.6 \\ \hline
\multirow{2}{*}{DLT (Slide Window)}
                &Best& \textbf{95.8} & \textbf{95.6} & 94.6 & 94.3 & \textbf{93.5} & \textbf{92.8} \\ 
                &Last& \textbf{95.4} & \textbf{95.3} & 94.0 & \textbf{93.9} & \textbf{93.1} & \textbf{92.2} \\ \hline
\bottomrule[1.5pt]
\end{tabular}
\end{center}
\vspace{-10pt}
\caption{Comparison with state-of-the-art methods in test accuracy ($\%$) on CIFAR-10 with different noise types and levels. We re-implement all methods under the same setting and the best results are boldfaced.}
\end{table*}
\subsection{Evaluating the effect of hard samples}
 As mentioned above, we suppose hard samples are always quite close to the decision boundary thus difficult to be classified by DNNs. To further provide some insights into what role hard samples play during the training process, we plan to conduct some experiments by involving some hard samples. However, it is very difficult to collect hard samples in real world. 
 Therefore, to evaluate the effect, we propose to generate two types of hard samples by using some techniques refer to image data augmentation~\cite{shorten2019survey} from different aspects:

\vspace{5pt}
\textbf{Samples with incomplete information}. By randomly erasing, cropping, rotation, resizing and affine transformation, we can generate hard samples by removing some features from original samples. 

\vspace{5pt}
\textbf{Samples suffered adversarial perturbation}. These samples are generated by adding some disturbances to origin samples. The adversarial examples of deep neural networks (DNNs) have attracted widespread attention~\cite{szegedy2014intriguing}. We propose to add a small number of adversarial perturbations to original samples and consider them as hard samples. Practically, this perturbation strategy is attacking against the well-trained model according to FGSM method~\cite{goodfellow2014explaining}. 
Formally, let $\epsilon$ be adversarial coefficient, $J(x,y)$ be the loss function. For each sample $x_i$, its generated adversarial sample $x_i^{adv}$ can be described as:
\begin{equation}
\begin{aligned}
x_i^{adv}=x_i + \epsilon \cdot sign(\nabla_{x} J(x_i,y_i))
\end{aligned}
\end{equation}

\section{Experiments}
We first conduct experiments in extensive settings and compare DLT to recent outstanding baselines. Besides, we empirically verify the proposed method for estimating noise rate by loss difference. Lastly, the effect of hard samples is also discussed in this section.
\begin{table*}[t]
\begin{center}
\renewcommand\arraystretch{1.1}
\begin{tabular}{p{3.5cm}<{\centering}p{1cm}<{\centering}|p{1cm}<{\centering}|p{1cm}<{\centering}|p{1cm}<{\centering}||p{1.0cm}<{\centering}|p{1cm}<{\centering}|p{1cm}<{\centering}}
\toprule[1.5pt]\hline
\multicolumn{2}{c|}{Noise Type}&\multicolumn{3}{c||}{Symmetric Noise} & \multicolumn{3}{c}{Asymmetric Noise}\\\hline
\multicolumn{2}{c|}{Methods / Noise Rate}&0.2& 0.4& 0.6& 0.2& 0.3& 0.4\\
\hline\hline
\multirow{2}{*}{Standard Cross-Entropy}
                &Best& 61.4 & 53.2 & 42.0 & 63.3 & 56.0 & 46.2 \\ 
                &Last& 57.5 & 41.2 & 24.4 & 59.6 & 51.8 & 42.6 \\ \hline
\multirow{2}{*}{APL (2020)}
                &Best& 71.2 & 66.7 & 54.0 & 70.1 & 66.4 & 55.5   \\ 
                &Last& 66.6 & 58.3 & 44.5 & 60.0 & 49.9 & 44.1   \\ \hline
\multirow{2}{*}{SCE (2019)}
                &Best& 60.1 & 53.7 & 42.8 & 60.2 & 55.6 & 45.8 \\ 
                &Last& 56.8 & 41.3 & 24.8 & 58.2 & 50.3 & 40.8 \\ \hline
\multirow{2}{*}{M-correction (2019)}
                &Best& 73.9 & 71.8 & 59.7 & --- & --- & --- \\ 
                &Last& 73.4 & 71.1 & 50.7 & --- & --- & --- \\ \hline
\multirow{2}{*}{DivideMix (2020)}
                &Best& 75.7 & 73.9 & 69.6 & 75.4 & 72.7 & 59.7 \\ 
                &Last& 74.9 & 73.0 & 69.1 & 74.9 & 72.0 & 51.2 \\ \hline
\multirow{2}{*}{DLT (Last Epoch)}
                &Best& 76.6 & 74.8 & 69.7 & \textbf{76.5} & 72.4 & 66.7 \\ 
                &Last& 75.5 & 73.8 & 69.0 & \textbf{75.1} & 71.3 & 65.5 \\ \hline
\multirow{2}{*}{DLT (Slide Window)}
                &Best& \textbf{77.1} & \textbf{75.2} & \textbf{70.1} & 76.1 & \textbf{72.8} & \textbf{68.0} \\ 
                &Last& \textbf{75.9} & \textbf{74.0} & \textbf{69.2} & 74.7 & \textbf{72.1} & \textbf{66.7} \\ \hline
\bottomrule[1.5pt]
\end{tabular}
\end{center}
\vspace{-10pt}
\caption{Comparison with state-of-the-art methods in test accuracy ($\%$) on CIFAR-100 with different noise types and levels. We re-implement all methods under the same setting and the best results are boldfaced.}
\end{table*}
\subsection{Experimental Setup}
\textbf{Datasets} To verify the superiority of our approach, we conduct experiments on two commonly used image classification datasets, namely CIFAR-10, CIFAR-100~\cite{krizhevsky2009learning}. Both of these two datasets contain 50,000 training and 10,000 test images, consisting of 32$\times$32 color images arranged in 10 and 100 classes, respectively. In addition, we also conduct experiments on Clothing1M~\cite{xiao2015learning}, which contains 14 classes with 1M real-world noisy training samples collected from online shopping websites. The labels are generated by the surrounding text of images and thus extremely noisy. Its overall noise rate is approximately 38$\%$.

\vspace{5pt}
\textbf{Noise Setting} Following previous works~\cite{arazo2019unsupervised,li2020dividemix,wang2019symmetric}, we generate two types of label noise: symmetric (uniform) noise and asymmetric (class-conditional) noise. Symmetric noisy labels are generated by randomly replacing the labels for a percentage of training data with all possible labels (i.e. the true label could be randomly maintained). While asymmetric noisy labels are only replaced by a specific set of classes (e.g. BIRD$\to$AIRPLANE, CAT$\leftrightarrow $DOG).

\vspace{5pt}
\textbf{Comparison Methods} We compare DLT with multiple baselines under the same noise settings. Here we introduce some of the most recent state-of-the-art methods. APL~\cite{ma2020normalized} combines two robust loss functions namely active loss and passive loss that mutually boost each other. SCE~\cite{wang2019symmetric} combines the Cross Entropy Loss with a noise robust counterpart named Reverse Cross Entropy (RCE). M-correction~\cite{arazo2019unsupervised} is the state-of-the-art loss correction approach which is specifically designed for symmetric noise, thus we only report its results under symmetric noise setting. DivideMix~\cite{li2020dividemix} is currently the state-of-the-art method for learning from noisy labels which leverages SSL techniques. The last two methods apply \textit{Mixup} augmentation.

\vspace{5pt}
\textbf{Implementation}
The implementation is based on PyTorch~\cite{paszke2019pytorch} and experiments were carried out with NVIDIA Tesla V100 GPU. We use an 18-layer PreAct Resnet~\cite{he2016deep} and train it using SGD with a momentum of 0.9, a weight decay of 0.0005, and a batch size of 128. The network is trained for 300 epochs. The warm up period is 10 epochs for CIFAR-10 and 30 epochs for CIFAR-100. We set the initial learning rate as 0.02 and reduce it by a factor of 10 after 150 epochs. For other hyperparameters of DLT, we set $T=0.5, K=2, s=200, \theta = 0.5$ and $ \lambda _{u} = \left \{ 0,25,150 \right \} $ under different noise rate. We re-implement all the comparison methods using the same network architecture, learning rate policy, and number of epochs with our proposed method. Note that DivideMix trains two networks simultaneously and uses the ensemble results in inference phase. For a fair comparison, we use the prediction from a single network. And we show that we can improve DLT by using the same averaging strategy at the inference phase in the following generalization experiments.

\begin{table}[t]
\begin{center}
\renewcommand\arraystretch{1.1}
\begin{tabular}{p{3.5cm}<{\centering}|p{2.5cm}<{\centering}}
\hline\hline
Method                 & Test Accuracy  \\ \hline
Standard Cross Entropy & 69.21     \\ \hline
F-correction (2017)    & 69.84     \\ \hline
M-correction (2019)    & 71.00     \\ \hline
Meta-Learning  (2019)  & 73.47     \\ \hline
DivideMix (2020)       & \textbf{74.76}     \\ \hline
DLT (Last Epoch)      & 74.04     \\ \hline
DLT (Slide Window)    & 73.58     \\ \hline
\hline
\end{tabular}
\end{center}
\vspace{-10pt}
\caption{Comparison with state-of-the-art methods in test accuracy ($\%$) on Clothing1M. Results of other methods are copied from the original papers.}
\end{table}

\begin{table*}[t]
\begin{center}
\renewcommand\arraystretch{1.1}
\begin{tabular}{m{3.5cm}<{\centering}m{1cm}<{\centering}|m{1cm}<{\centering}|m{1cm}<{\centering}|m{1cm}<{\centering}||m{1.0cm}<{\centering}|m{1cm}<{\centering}|m{1cm}<{\centering}}
\hline\hline
\multicolumn{2}{c|}{Noise Type}& \multicolumn{3}{c||}{CIFAR-10} & \multicolumn{3}{c}{CIFAR-100} \\ \hline
\multicolumn{2}{c|}{Methods / Noise Rate}&0.2&0.5&0.8&0.2& 0.5& 0.8 \\ \hline\hline
\multirow{2}{*}{DivideMix}
& Best & 96.1 & 94.6 &\textbf{93.2}& 77.3 & 74.6 & 60.2 \\ 
& Last & 95.7 & 94.4 & 92.9 & 76.9 & 74.2 & 59.6 \\ \hline
\multirow{2}{*}{DLT*}& Best &\textbf{96.5}&\textbf{95.5} &\textbf{93.2} & \textbf{80.0} & \textbf{75.9} & \textbf{60.8}  \\  
& Last & \textbf{96.1} & \textbf{95.2}  &  \textbf{93.0} & \textbf{79.2} & \textbf{75.2} & \textbf{60.4}  \\ \hline
\hline                                      
\end{tabular}
\end{center}
\vspace{-10pt}
\caption{Comparison with DivideMix in test accuracy ($\%$) on CIFAR-10 and CIFAR-100 with different noise rates under symmetric label noise.}
\end{table*}

\begin{table*}[t]
\begin{center}
\begin{tabular}{m{3.2cm}<{\centering}|m{0.7cm}<{\centering}|m{0.7cm}<{\centering}|m{0.7cm}<{\centering}|m{0.7cm}<{\centering}|m{0.7cm}<{\centering}||m{0.7cm}<{\centering}|m{0.7cm}<{\centering}|m{0.7cm}<{\centering}|m{0.7cm}<{\centering}|m{0.7cm}<{\centering}}
\hline\hline
                     & \multicolumn{5}{c||}{CIFAR-10} & \multicolumn{5}{c}{CIFAR-100} \\ \hline
Real Noise Rate      & 20   & 30   & 40   & 50   & 60   & 20   & 30   & 40   & 50   & 60  \\ \hline
Estimated Noise Rate & 21.6 & 31.8 & 40.9 & 49.5 & 59.1 & 25.3 & 31.5 & 39.8 & 47.8 & 56.6     \\ \hline\hline
\end{tabular}
\end{center}
\vspace{-10pt}
\caption{Noise rate ($\%$) estimation of our proposed method on CIFAR-10 and CIFAR-100.}
\end{table*}

\subsection{Experimental Results}

\textbf{Experiments on CIFAR-10/100}
Table 1 and 2 present the results on benchmark datasets CIFAR-10 and CIFAR-100  with different noise types and levels. We report both the best test accuracy across all epochs and the averaged test accuracy over the last 10 epochs for each case. DLT with slide window strategy outperforms existing state-of-the-art methods across all noise rates with different noisy types, especially on CIFAR-10 under 0.4 and 0.6 symmetric noise. Besides, merely on CIFAR-10 under 0.4 asymmetric noise and CIFAR-100 under 0.3 asymmetric noise, DLT with last epoch strategy falls short of the state-of-the-art method DivideMix. Note that the performance of slide window strategy is slightly superior to the last epoch strategy in most cases. We postulate this is because the loss values of the last batches can reflect more real-time information from DNNs.

\vspace{5pt}
\textbf{Experiments on real-world dataset}
We have seen that DLT achieves excellent performance on datasets with manually corrupted noisy labels. Next, we conduct another experiment on a real-world large-scale noisy dataset: Clothing1M, and the comparison results are reported in Table 3. In this case, we use ResNet-50 with ImageNet pre-trained weights. The test accuracy of DLT is lower than that of the state-of-the-art method about $1\%$. We think this limitation is caused by the instance-dependent noisy labels in Clothing1M dataset and our proposed method is not robust to this type of noise. Furthermore, the result of DivideMix is based on the ensemble of two networks.

\vspace{5pt}
\textbf{Generality of the proposed approach}
To demonstrate that DLT has an advantage in generalization and flexibility, we apply it to DivideMix~\cite{li2020dividemix}. Note that we use the same network, hyperparameters (except we fix $w_b$ as 0.5), and learning rate policy as DivideMix. As shown in Table 4, our noisy label detection method consistently outperforms DivideMix, even when the noise rate is high. Note that combined with the schemes reported in DivideMix, DLT achieves great improvement on CIFAR-100 under 0.2 and 0.5 symmetric noise. This indicates that DLT is of excellent generality and can still be effective when applied to other methods.

\vspace{5pt}
\textbf{Experiments on noise rate estimating}
As we described in Section 3.2, the noise rate of the original dataset can be estimated by our proposed approach based on loss difference between the early and late training stages. Specifically in our experiments, we simply choose the $30_{th}$ epoch as the early training stage and the $300_{th}$ (last) epoch as the late training stage respectively. Moreover, we obtain the loss difference value of each sample from these two selected epochs. Then we estimate the noise rate using Eq. (10). As shown in Table 5, we can get highly accurate estimations on CIFAR-10 and CIFAR-100 across the noise rate from 0.2 to 0.6 under symmetric noise, which verifies the effectiveness of our method.


\vspace{5pt}
\textbf{Discussion about hard samples}
We experimentally investigate the effect of hard samples in the DNN training from noisy labels. In order to simulate the real-world situations, we generate hard samples according to the data augmentation strategies introduced in Section 3.3. In our experiments, we manually add the same proportion of hard samples as the original clean ones into datasets. As shown in Figure 7, the loss value changing tendency of hard samples behaves similarly to that of original clean samples, but has significant differences from that of noisy samples. According to the diverse loss values, our method can divide the hard samples into the clean set rather than the noisy set. This verifies the effectiveness of DLT although existing some indistinct samples.

\section{Conclusion}
In this paper, we propose a simple but effective method DLT, which is based on dynamic loss thresholds for learning from noisy labels. DLT takes advantage of different loss value distributions of clean and noisy samples and leverages semi-supervised learning techniques to exploit the mislabeled data. Along with gradually decreasing loss thresholds and slide window strategy, DLT achieves state-of-the-art performance. Besides, we propose a noise rate estimation method based on loss difference and achieve considerable results both on CIFAR-10 and CIFAR-100. In the end, we experimentally verify the effectiveness of our noisy label detection method in the scenario that some hard samples are contained in training data.

\begin{figure}[t]
\begin{center}
\includegraphics[width=0.9\linewidth]{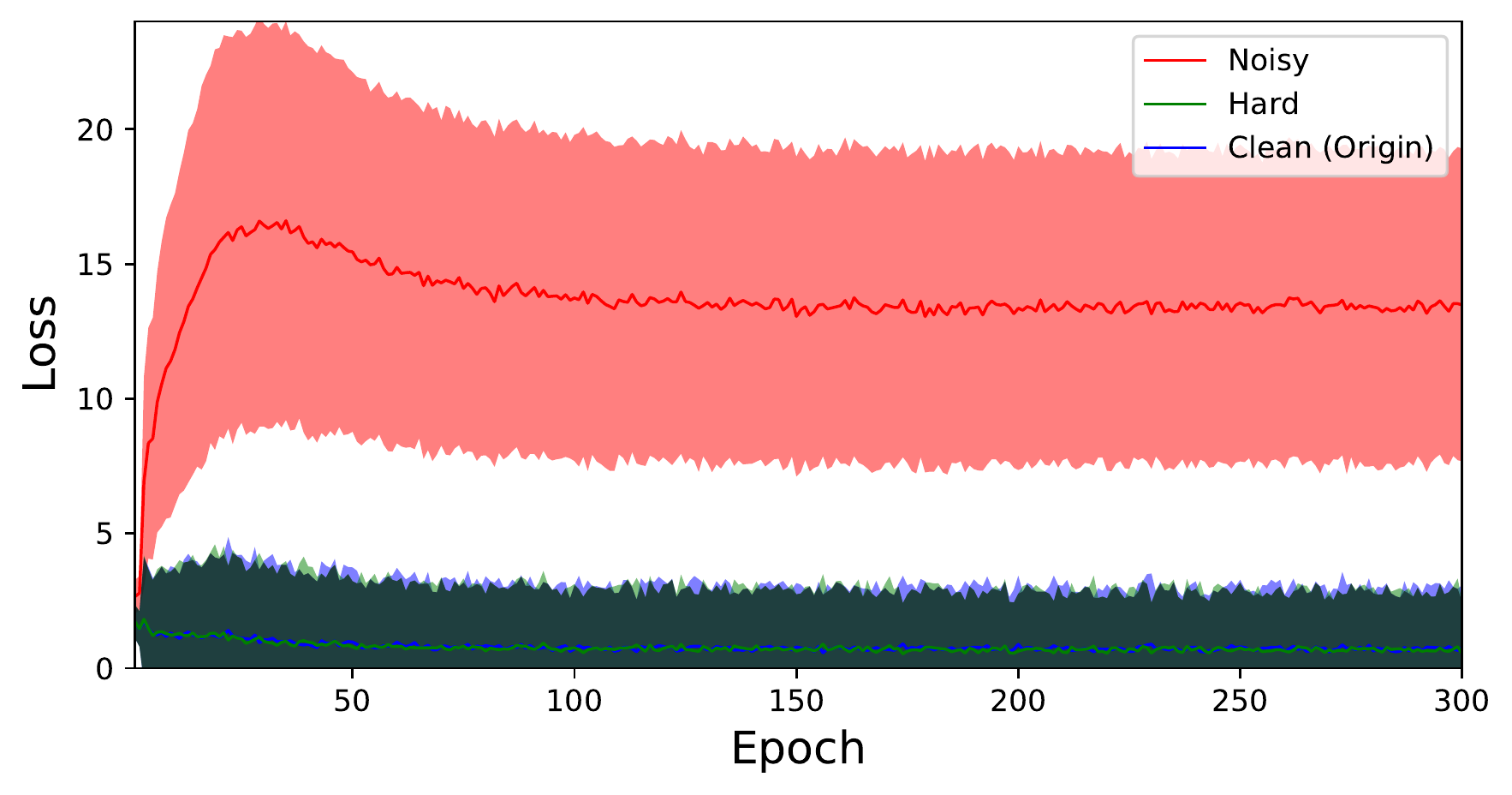}
\end{center}
\vspace{-5pt}
\caption{Loss tendencies of noisy samples, hard samples and original clean samples.}
\label{fig:long}
\label{fig:onecol}
\end{figure}

{\small
\bibliographystyle{ieee_fullname}
\bibliography{ref}
}

\end{document}